\documentclass[pdflatex,sn-basic,hidelinks]{sn-jnl}
\usepackage{anyfontsize}

\usepackage{graphicx}%
\usepackage{multirow}%
\usepackage{amsmath,amssymb,amsfonts}%
\usepackage{amsthm}%
\usepackage{mathrsfs}%
\usepackage[title]{appendix}%
\usepackage{xcolor}%
\usepackage{textcomp}%
\usepackage{manyfoot}%
\usepackage{booktabs}%
\usepackage{algorithm}%
\usepackage{algorithmicx}%
\usepackage{algpseudocode}%
\usepackage{listings}%
\usepackage{gensymb}
\usepackage{hyperref}

\theoremstyle{thmstyleone}%
\theoremstyle{thmstyletwo}%

\theoremstyle{thmstylethree}%

\hypersetup{colorlinks=false}
\begin{document}

\title[Article Title]{Identifying environmental factors associated with tetrodotoxin contamination in bivalve mollusks using eXplainable AI}


\author*{\fnm{M.C.} \sur{Schoppema}}\email{thijs.schoppema@wur.nl}
\author{\fnm{B.H.M.} \spfx{van der} \sur{Velden}}
\author{\fnm{A.} \sur{Hürriyetoğlu}}
\author{\fnm{M.D.} \sur{Klijnstra}}
\author{\fnm{E.J.} \sur{Faassen}}
\author{\fnm{A.} \sur{Gerssen}}
\author{\fnm{H.J.} \spfx{van der} \sur{Fels-Klerx}}

\affil{\orgname{Wageningen Food Safety Research}, \orgaddress{\street{Akkermaalsbos 2}, \city{Wageningen}, \postcode{6708WB}, \country{Netherlands}}}

\abstract{Since 2012, tetrodotoxin (TTX) has been found in seafoods such as bivalve mollusks in temperate European waters. TTX contamination leads to food safety risks and economic losses, making early prediction of TTX contamination vital to the food industry and competent authorities. Recent studies have pointed to shallow habitats and water temperature as main drivers to TTX contamination in bivalve mollusks. However, the temporal relationships between abiotic factors, biotic factors, and TTX contamination remain unexplored.

We have developed an explainable, deep learning-based model to predict TTX contamination in the Dutch Zeeland estuary. Inputs for the model were meteorological and hydrological features; output was the presence or absence of TTX contamination. 

Results showed that the time of sunrise, time of sunset, global radiation, water temperature, and chloride concentration contributed most to TTX contamination. Thus, the effective number of sun hours, represented by day length and global radiation, was an important driver for tetrodotoxin contamination in bivalve mollusks. 

To conclude, our explainable deep learning model identified the aforementioned environmental factors (number of sun hours, global radiation, water temperature, and water chloride concentration) to be associated with tetrodotoxin contamination in bivalve mollusks; making our approach a valuable tool to mitigate marine toxin risks for food industry and competent authorities.
}



\maketitle

\section{Introduction}\label{intro}
Marine biotoxins pose a serious threat to human health, causing severe illnesses such as paralytic, amnesic, diarrheic, and neurotoxic poisoning \citep{Visciano2016}. These toxins not only impact human health but also lead to severe economic losses and supply chain disruptions in the food industry \citep{Kouakou2019}. These impacts, losses, and disruptions make the early prediction of seafood contamination vital for reducing food safety hazards and associated economic losses.

One group of relevant biotoxins in seafood is tetrodotoxins (TTXs). TTXs are potent neurotoxins that, upon human consumption, can result in paralysis, and, at elevated levels, respiratory failure \citep{Katikou2022}. TTXs are generally associated with seafood from tropical waters. However, since 2012, TTX has been regularly detected in bivalve mollusks from temperate European waters, including waters of Greece \citep{Vlamis2015}, England \citep{Turner2015}, the Netherlands \citep{Gerssen2018}, Portugal \citep{Cruz2022}, Spain \citep{Leo2018}, Italy \citep{Bordin2021, DellAversano2019}, and France \citep{Hort2020}. In 2017, the European Food Safety Authority (EFSA) published an opinion on the human health risks related to the presence of TTX contamination in bivalve mollusks \citep{Knutsen2017}. EFSA stated the importance of further data collection focused on sources and critical factors leading to TTX contamination. 

Following the EFSA publication, additional samples of TTX contaminated bivalve mollusks were detected by national authorities. These additional data provided an improved basis to link environmental factors with TTX contamination. The following factors have been hypothesized to correlate with TTX contamination in bivalve mollusks: shallow estuarine waters, water pH value, refreshment rates, and water temperature \citep{Gerssen2018, Turner2017}. Specifically, the temperature range contributing to the presence of TTX was defined as $15-20\celsius$ \citep{Dhanji-Rapkova2023}. Despite the identification of these factors, the exact mechanism underlying TTX contamination in bivalve mollusks remains unknown \citep{Katikou2022}. The identified environmental factors are only a subset of all impactful factors. In addition, the origin of TTX contamination is also unknown.

The aforementioned research on TTX performed correlation analyses, which cannot sufficiently describe the complex (temporal) relations underlying TTX contamination in bivalve mollusks. Artificial intelligence (AI) has the potential to model these complex relations and to provide additional hints towards the mechanism underlying TTX contamination in bivalve mollusks. Furthermore, AI can serve as a tool for the early prediction of contamination, potentially decreasing response time. From the AI technologies, deep learning has specifically shown the potential to predict the presence of contaminants other than TTX in seafood \citep{Cruz2022, Ma2024, Marzidovek2024, Tavares2023}.

TTX can accumulate over time in bivalve mollusks \citep{Dhanji-Rapkova2023}, an accumulation which, as previously discussed, seems dependent on several environmental factors. These factors form a complex interplay which makes deep learning techniques designed for time series data, specifically a long short-term memory (LSTM) model \citep{Hochreiter1997}, the preferred choice. An LSTM is able to take long and short-term information into account. In addition to its specialization with time series data, such an algorithm can also simultaneously analyze several environmental factors. Furthermore, when combined with eXplainable AI (XAI) methods, it becomes feasible to analyze the LSTM’s reasoning and extract the importance of each environmental factor. 

The aim of this study is to predict the presence of TTX-contaminated bivalve mollusks using explainable deep learning to understand which conditions drive TTX contamination. An approach which detects marine toxin contamination and explains its possible drivers would be vital for competent authorities and food industries to mitigate marine toxin risks.

\section{Results}\label{res}
\subsection{Data}
Data from the official Dutch shellfish monitoring program contained analytical results of 3,143 samples from 2016-2023. TTX concentrations above the Limit of Detection (LOD) were found in 331 samples (11\%), and TTX was not detected in 2,812 samples (89\%). After removing duplicates, as described in section \ref{clean}, 1,156 samples remained, of which 222 were with TTX concentrations above the LOD (19\%), and 934 were with TTX concentrations below LOD (81\%). Supplementary \autoref{REGIONS} shows a breakdown of samples by region. Of the 222 samples with TTX above LOD, 75 (34\%) had a TTX concentration above the Action Limit (AL, 22 \micro g TTX/kg) and 44 (20\%) above the Legal Limit (LL, 44 \micro g TTX/kg). The AL is a minimum concentration of TTX, after which, upon detection, the monitoring strategy intensifies to weekly sampling.

Meteorological data were acquired from the Royal Netherlands Meteorological Institute (KNMI) \citep{KoninklijkNederlandsMeteorologischInstituut2024}. Hydrological data were acquired from Rijkswaterstaat (RWS) \citep{Rijkswaterstaat2024}. Supplementary \autoref{DATASIZE} shows the 35-day average characteristics of the meteorological and hydrological features. The train set (2016-2021) had 47 samples with TTX detected, the validation set (2022) had 13 samples with TTX, and the test set (2023) had 15 samples with TTX. Supplementary \autoref{DATADIST} shows specific distributions of the train, validation, and test sets.

\subsection{Deep learning performance}
\begin{figure}[t]
\centering
\includegraphics[width=0.75\textwidth]{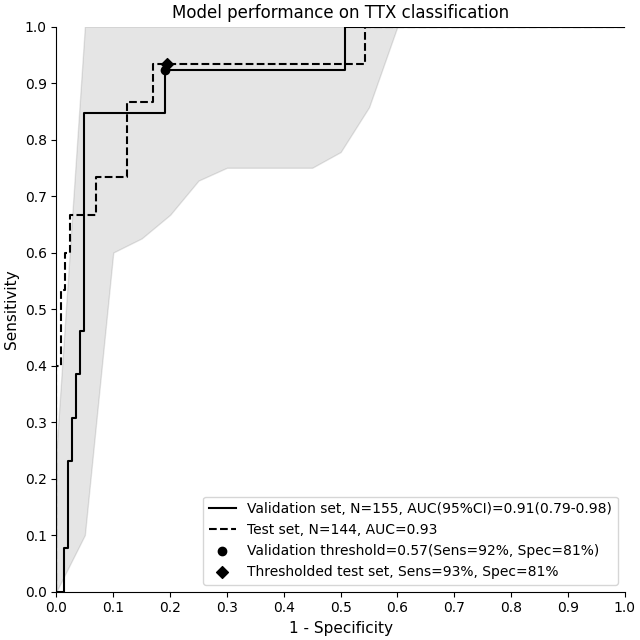}
\caption{
Receiver operating characteristics of above Action Limit (AL) classification by our long short-term memory (LSTM) method on the validation set (N=154, area under the curve (AUC)=0.91, solid line), and test set (N=144, AUC=0.93, dotted line). The 95\% Confidence Interval (CI) (AUC=0.79-0.98, shaded area) was obtained by bootstrapping the validation set 10,000 times. The performance on the test set shows similar results to the validation set. There appears to be one TTX positive  sample in the test set, which the model fails to correctly classify. Finally, at a threshold of 0.57, selected based on a validation sensitivity >90\%, the test set obtained a sensitivity (Sens) of 93\% and a specificity (Spec) of 81\% (diamond icon).
}\label{ROCAL}
\end{figure}

The LSTM had an area under the curve (AUC) of 0.91 (95\% confidence interval, 95\% CI: 0.79-0.98) in the validation set (\autoref{ROCAL}). At a sensitivity of 90\%, the validation specificity was 81\% (95\% CI: 44\%-96\%) with an associated threshold of 0.57. The test AUC was 0.93, which falls within the bootstrapped confidence interval of the validation set. At a test sensitivity of 90\%, the test specificity was 83\%, which also falls within the bootstrapped confidence interval of the validation set (\autoref{ROCAL}). At a threshold of 0.57, the test specificity was 81\% in comparison to a validation specificity of 81\%, and test sensitivity was 93\% in comparison to a validation sensitivity of 92\%.

\begin{figure}[t]
\centering
\includegraphics[width=0.9\textwidth]{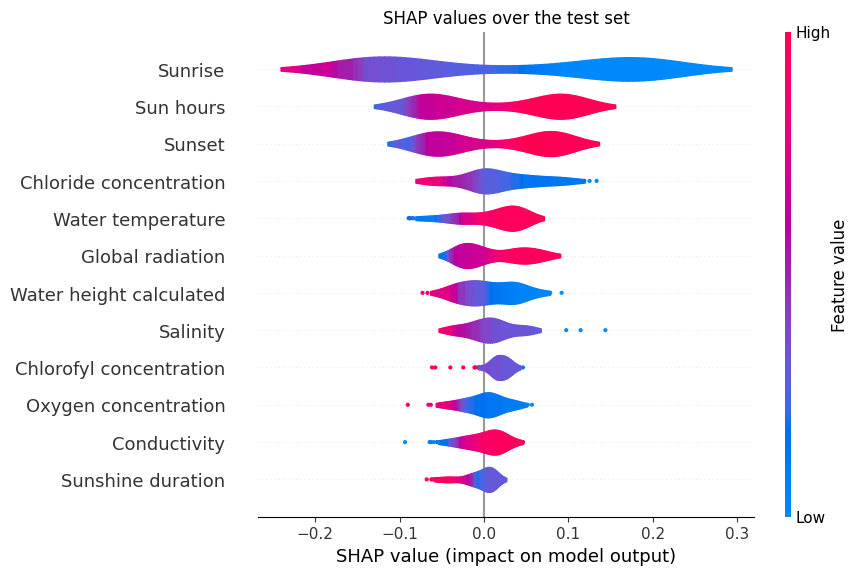}
\caption{
Explainable AI (XAI) shows, with SHaply Additive exPlanations (SHAP), the global explanations for the long short-term memory model. The most important features are time of sunrise, time of sunset, sun hours, chloride concentration, global radiation, and water temperature. Each of these features were shown to be significant ($p < 0.05$) for the model’s prediction, with the exception being water temperature ($p = 0.10$).
}\label{SHAPS}
\end{figure}

\subsection{Explainable AI}
\begin{figure}[H]
\centering
\includegraphics[width=0.6\textwidth]{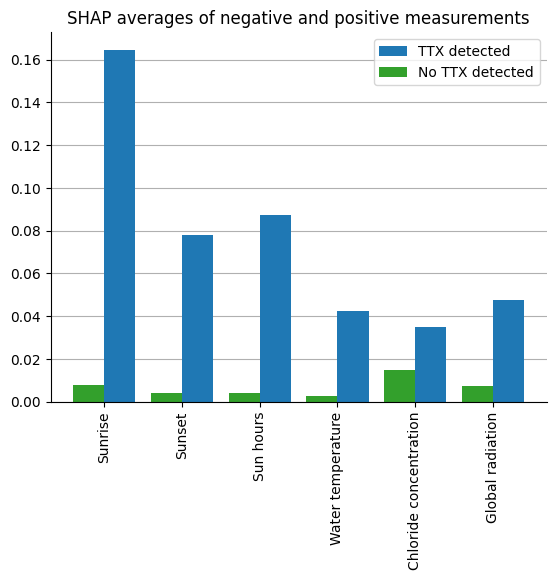}
\caption{
Difference in average SHAP values between samples with a TTX concentration below the action limit (green bars, left) and above the action limit (blue bars, right) for the six most important features according to our XAI method. Each of these features was shown to be significant ($p < 0.05$) for the model’s prediction, except water temperature ($p = 0.10$).
}\label{DIFFSHAPS}
\end{figure}

The XAI method SHapley Additive exPlanations (SHAP) showed that the most important features included seasonal related attributes, as shown by time of sunrise, time of sunset, and number of sun hours (\autoref{SHAPS}). The most important hydroclimatic features that correlate with TTX contamination were a low chloride concentration, high global radiation, and high water temperature. Other contributing factors to TTX contamination were water height, salinity, conductivity, and oxygen concentration.
The global explanations of the most important features show a significant ($p < 0.05$) difference between analytical results that were positive and negative for TTX contamination (\autoref{DIFFSHAPS}), except for water temperature ($p = 0.10$).

\begin{figure}[H]
\centering
\includegraphics[width=0.9\textwidth]{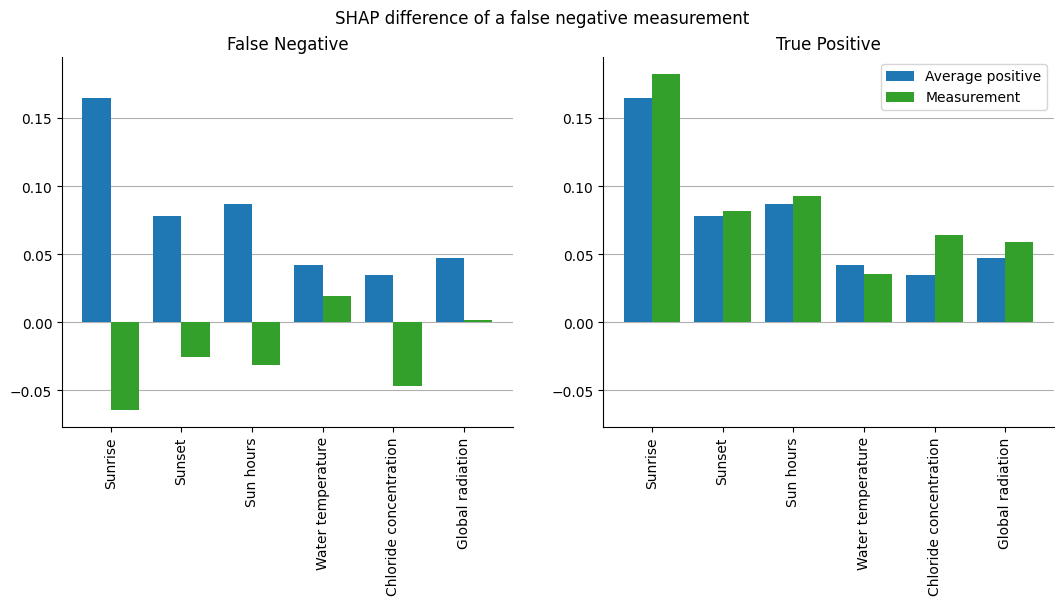}
\caption{
Left: The SHAP values of the false negative identified in \autoref{ROCAL}. Right: a correctly classified true positive sample is shown. Besides each SHAP value, the average SHAP values of all TTX positive samples are shown. The false negative sample diverges in prediction direction on all features, except water temperature.
}\label{FNTPSHAP}
\end{figure}

One sample was positive for TTX contamination, but the model predicted the sample to have a low probability (probability = 0.21, average = 0.58) for being positive (i.e., a false negative) (\autoref{ROCAL}). Local explanations of this false negative prediction \autoref{FNTPSHAP} showed that its SHAP values follow a different pattern when compared to a true positive case. The false negative diverges on all features apart from water temperature, whereas for true positives do the features do not diverge (\autoref{FNTPSHAP}).

\subsection{Sensitivity analyses}
\begin{figure}[H]
\centering
\includegraphics[width=0.75\textwidth]{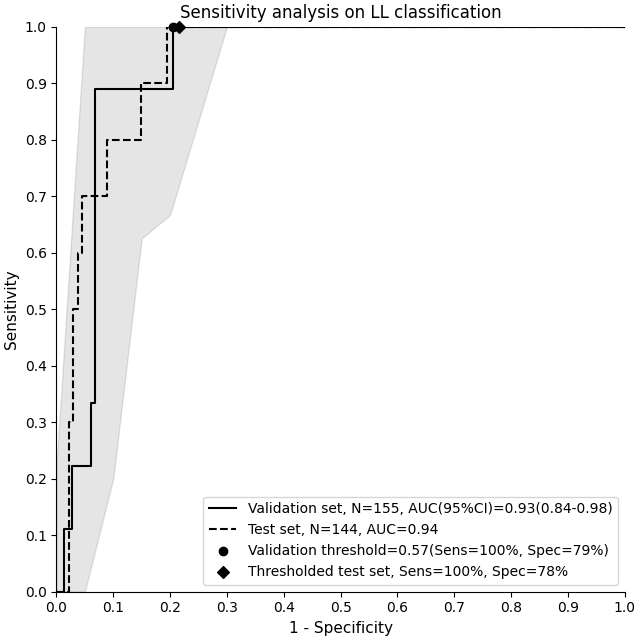}
\caption{
Receiver operating characteristics of above Legal Limit (LL) classification by our long short-term memory (LSTM) method on the validation set (N=154, area under the curve (AUC)=0.93, solid line), and the test set (N=144, AUC=0.94, dotted line). The 95\% Confidence Interval (CI) (AUC=0.84-0.98, shaded area) was obtained by bootstrapping the validation set 10,000 times. As with the main analysis (\autoref{ROCAL}), the sensitivity analysis showed a test set performance similar to the validation set. Finally, at a threshold of 0.57, selected based on a validation sensitivity $>$90\%, the test set obtained a sensitivity (Sens) of 100\% and specificity (spec) of 78\% (diamond icon).
}\label{LLROC}
\end{figure}

The sensitivity analyses focused on the Legal Limit (LL) and showed similar results as the Action Limit (AL)-based test set. When testing the LSTM on detecting analytical results above the LL, this resulted in an AUC of 0.94 (validation 95\% CI 0.84-0.98), with a specificity of 81\% (validation 95\% CI 73\%-96\%) at a sensitivity of 90\% (\autoref{LLROC}). At a threshold of 0.57, the  specificity in the test set was 78\%, and the sensitivity in the test set was 100\%. 

To measure the impact of the region Eastern Scheldt Middle and its data imputation method, we tested the LSTM on additional test sets without Eastern Scheldt Middle. Both the AL and LL test set without Eastern Scheldt Middle had an AUC of 0.92.

\section{Discussion}\label{disc}
We showed that time series deep learning can predict the presence of TTX in bivalve mollusks in Zeeland. XAI provided insights into the features most associated with both positive and negative predicted TTX cases, thereby enhancing scientific understanding into environmental factors driving TTX contamination in bivalve mollusks.

Our method predicted TTX contamination above the AL with an AUC of 0.93 and achieved a high specificity of 83\% at a sensitivity of 90\%. In addition to the prediction of TTX contamination above the AL, the model was also able to predict TTX contamination above the LL. Our method can thus adapt to different thresholds for contamination and needs of public and private monitoring programs. We emphasize that our deep learning method was trained on data from consecutive years and tested on data from the last independent year. This deliberate split in data yields insights into potential prospective application of the model and its performance, thereby ensuring generalizability for future analytical results. 

Our XAI method identified six major temporal features associated with the presence of TTX contamination: time of sunset, time of sunrise, daily number of sun hours, global radiation, water temperature, and water chloride concentration. The features time of sunset, time of sunrise, daily number of sun hours, and global radiation show the important role solar intensity has on TTX contamination. Currently, an exogenous origin such as bacteria or algae is the most accredited hypothesis of TTX contamination of bivalve mollusks \citep{Bacchiocchi2022, Rodrguez2017, Vlamis2015}. The impact of solar intensity on TTX contamination can be explained by the suspected exogenous origin: prior studies show that the activity of most bacteria, phytoplankton, and worms is positively correlated with solar intensity \citep{Edwards2015}. Furthermore, our findings also confirm the suspected importance of shallow waters \citep{Turner2015}, since lower water depth results in a higher solar intensity due to the absorption of light by water. 

Unlike solar intensity, chloride concentration correlates negatively with the presence of TTX. Additionally, while not significant, the model also hints to other factors such as salinity and conductivity to impact TTX contamination. The importance of values such as chloride concentration, conductivity, and salinity, together with the dependence on temperature, could explain why TTX contamination mainly appears during late spring and early summer, compared to similar day lengths in the second half of the year, which are not associated with TTX contamination.  It should be noted that while the chloride concentration and salinity may impact TTX accumulation, their importance might also be explained by a confounder, since chloride concentration is affected by changing evaporation and precipitation rates, as well as by hydrological interventions such as the opening of sluices \citep{Augustijn2011}. We also observe the previously found relationship between increased water temperature and TTX contamination \citep{Antonelli2023, Bacchiocchi2022, Dhanji-Rapkova2023, Gerssen2018, Turner2017}, though the relationship is not significant ($p = 0.10$) when taking temporal aspects and additional features into account. 

In addition to the identified impactful features, the XAI method also showed the differences between false negative and true positive analytical samples. Local explanations showed why a TTX contaminated sample was not predicted TTX positive by the model, as the sample differentiated on most features from the average positive sample. These local explanations provide further insights for end users when evaluating such an AI system and, by extension, enhancing trust in the AI system. 

Prior studies have used AI, including deep learning, and available data to predict the presence of other contaminants than TTX in seafood. \cite{Cruz2022} used remote sensing and hydrological data with artificial neural networks (ANN) to forecast the presence of diarrhetic shellfish poisoning (DSP) toxins in bivalve mollusks along the Portuguese coast. \cite{Tavares2023} continued this work by combining remote sensing data with autoencoders and ANNs. \cite{Ma2024} used a LSTM model to forecast marine biotoxin concentrations in California using a combination of meteorological and hydrological data as input. Although these studies all incorporated AI to predict the presence of contaminants in seafood, none have considered the temporal aspect in combination with deep learning and XAI. A pitfall of deep learning methods is that they are inherently uninterpretable and often likened to a black box \citep{vanderVelden2023}. This characteristic makes it difficult to directly extract any meaningful knowledge from deep learning models. XAI enables us to analyze a deep learning model’s reasoning and thus provides insight in the black box. In our study, we have shown the importance of combining temporal deep learning with XAI. First, we provided the LSTM model with temporal information, enabling the model to learn temporal dependencies. Second, we used XAI to discover important temporal correlative features.

Our study has some limitations. First, there were missing hydrological measurements. To mitigate the effects of resulting missing data, we used imputation techniques. For irregular missing data, we used k-nearest neighbors imputation (kNN) \citep{Cover1967}; for monthly sampled data, we used a forward fill. For the region Eastern Scheldt Middle, we had to take the average values of the hydrological data from the neighboring regions due to a limited availability of hydrological measurements. This approach proved valid, since our model was robust for the considered regions, i.e., we did not find classification differences (at most 0.01 AUC on AL) between the results with and without the Eastern Scheldt Middle. Second, the dataset contained a relatively limited amount (75) of positive samples above the AL. It should be noted that this dataset was the result of an eight-year-long monitoring plan and is therefore very comprehensive. Nonetheless, it would be of interest to combine datasets from different countries to improve the modeling and increase statistical power. 

Our method facilitates decision-making in the public and private monitoring of TTX in bivalve mollusks. The AI model can predict the likelihood of TTX contamination. This prediction can be used to adapt the monitoring strategy, i.e., at high probabilities, to intensify the monitoring. Our method focused on TTX contamination in bivalve mollusks in one area of the Dutch coastal zone. The method can be extended to different contaminants, seafoods, and regions with available hydrological and meteorological measurements, in addition to contaminant monitoring results. The prediction of multiple contaminants at the same time could also be performed with a multi-task deep learning method, which can increase statistical power and thus provide an even more robust classification \citep{Crawshaw2020}. Additionally, the found importance of hydrological features by our method - such as chloride concentration and salinity - for TTX contamination in bivalve mollusks should encourage TTX monitoring strategies to include hydrological measurements. Specifically, with the identified importance of solar intensity on TTX contamination, measuring the clarity of water at growth sites might be beneficial for further research \citep{Rttgers2014}. Finally, it is important to note that our XAI method identifies correlative relationships. Expanding TTX monitoring with hydrological and meteorological measurements could therefore support efforts to further uncover the causative factors of TTX contamination.

To conclude, we showed that we could predict the presence of TTX contaminated bivalve mollusks using temporal deep learning-based AI. We provided explanations on how the AI reached its predictions. With this approach, we identified the impact of seasonal and hydroclimatic variables on TTX contamination in bivalve mollusks showing the importance of solar intensity, through descriptors such as length of day and global radiation, as a potential driver for TTX contamination. These findings make our approach a valuable tool for competent authorities and food industry to minimize marine toxin risks.

\section{Methods}\label{sec11}
In this study, we created an AI model to predict the presence of TTX contaminated bivalve mollusks. Inputs to the AI model were hydrological and meteorological data, henceforth called features, while output was presence or absence of detectable TTX levels in bivalve mollusks. All input was normalized, and missing values were imputed. The data were split by year for model training, validation, and testing ensuring a fair separation and an independent test set that mimics prospective validation \citep{vanMeer2025}. The model’s performance was evaluated by the AUC \citep{Metz1978}. Finally, after model creation, we used SHapely Additive exPlanations (SHAP) to provide insight into the model’s predictions \citep{Lundberg2017}. The following sections describe the creation and analysis of the AI model in more detail. 

\subsection{Data} 
\subsubsection{Monitoring data}\label{clean}
The analytical TTX contamination results, henceforth called samples, originated from the official Dutch shellfish monitoring program. Samples were collected from bivalve mollusk growth beds in Zeeland, which is part of the Dutch coastal area, in the period 2016-2023. The sample collection was performed monthly from October-May and weekly from June-October, except for weeks 24-28, when sampling was performed twice a week. The used TTX analysis protocol has been described in \cite{Alkassar2024}. In short, the analyses for TTX were performed using liquid chromatography-tandem mass spectrometry (LC-MS/MS). Chromatographic separation was carried out on a ZORBAX bonus RP RRHD 2.1 x 100 mm, 1.8 \micro m (Agilent Technologies, Santa Clara, CA, USA). The LC-MS/MS analyses were carried out with a Waters Acquity UPLC coupled to a Waters Xevo TQ-S tandem mass spectrometer (Waters, Milford, MA, USA). Limit of Detection (LOD) was 10 \micro g TTX/kg, Limit of Quantification was 20 \micro g TTX/kg, a concentration of 22 \micro g TTX/kg was the AL, and a concentration of 44 \micro g TTX/kg was the national LL \citep{Staatscourant2017, Staatscourant2022}. The AL is a minimum concentration of TTX, which, after detection, results in an intensified monitoring strategy of weekly sampling.

Each sample came from one of six subregions in Zeeland; the location of subregions is shown in supplementary \autoref{REGIONS}. When multiple samples on one day originated from the same subregion, distinguishing between them was not possible, as attempts to do so introduced ambiguity in the AI model. Therefore, in these cases, the sample with the highest TTX level was used; these samples are henceforth referred to as duplicates. This approach has minimal effect on the modeling, as the AI model only performs a binary classification.

\subsubsection{Meteorological and hydrological features}
Meteorological features were acquired from the data platform of the KNMI \citep{KoninklijkNederlandsMeteorologischInstituut2024} for the measurement station at Vlissingen (station number 310) and included daily mean temperature (in 0.1 \celsius), maximum temperature (in 0.1 \celsius), minimum temperature (in 0.1 \celsius), sunshine duration (in 0.1 hours), global radiation (in J/cm\textsuperscript{2}), average wind speed (in 0.1 m/s), average wind direction, precipitation duration (in 0.1 hours), and precipitation (in 0.1 mm). The measurement station at Vlissingen was the only homogenized station in Zeeland in near proximity of the shellfish production sites. 
Hydrological features were acquired from the data platform of RWS \citep{Rijkswaterstaat2024}, a part of the Ministry of Infrastructure and Water Management of the Netherlands, and included oxygen concentration, oxygen saturation, chlorophyll concentration, chloride concentration, chlorosity, pheophytin concentration, pH value, air pressure, (calculated) water height, water temperature, wind direction, wind speed, conductivity, and salinity. These features were sampled from multiple measurement stations across Zeeland, focusing on the Eastern Scheldt, Lake Veere, and Lake Grevelingen. The measurement stations have been appointed to different sub-regions, as shown in supplementary \autoref{REGIONS}. For sub-regions in which multiple measurements of one feature are taken on a given day, we used the mean value over the available stations.

\subsubsection{Data pre-processing \& normalization}
Outliers from the hydrological features, of which measurement errors, were removed by calculating the 2.5th and 97.5th percentile for each feature. With the percentiles, we removed hydrological features which fell outside the 2.5-97.5 percentile range. Since RWS does not measure all hydrological features on a day-by-day basis, missing hydrological features were imputed using weighted kNN \citep{Cover1967} from scikit-learn \citep{Pedregosa2011}. As the TTX monitoring plan follows (mostly) a (bi)-weekly pattern, we set the number of neighbors to seven. We used the kNN imputation for hydrological features with infrequent missing values. When hydrological features were only taken once a month (e.g., oxygen concentration), we imputed 30 days with a forward fill. In case a region had no values for a given hydrological feature, the average feature from neighboring regions was imputed. As typically carried out in deep learning, we min-max normalized meteorological and hydrological features between 0 and 1. 

\subsection{Deep learning}
We used an LSTM model to predict TTX contamination in bivalve mollusks \citep{Hochreiter1997}. We chose an LSTM because it excels in processing temporal data. Unlike other methods, such as the standard recurrent neural network, the LSTM is able to process and recognize long-term relationships in the data.
Input to the LSTM were the hydrological and meteorological features; output was whether a sample was positive or negative for TTX. An analytical result was considered positive if above the AL of 22 \micro g TTX/kg. We provided the LSTM model with 5 weeks (35 days) of hydrological and meteorological features prior to each analytical result. A period of 5 weeks has been chosen because earlier studies indicated that temperatures of at least three weeks prior were indicative of TTX contamination \citep{Hort2020}. We split all analytical results on sampling year into a train set (2016-2021), validation set (2022), and holdout test set (2023). Splitting the samples by year prevents data leakage, ensuring a fair evaluation \citep{vanMeer2025}.

The LSTM consisted of an LSTM module and classification head. The classification head consisted of three linear layers (128, 64, and 2 neurons). We used layer normalization (Ba et al., 2016) after each LSTM layer. We applied batch normalization \citep{Ioffe2015} after the first linear layer. Normalization layers were used with dropout \citep{Srivastava2014} to prevent overfitting. Each layer used a ReLU activation function \citep{Agarap2019}, except the last layer, which had no activation function due to the used loss function. However, during inference, the last layer used softmax activation.

We used the AdamW optimizer \citep{Loshchilov2019} with a weighted cross-entropy loss. The cross-entropy loss was weighted on the train-class distribution to mitigate class imbalance. supplementary \autoref{PARAMS} shows the values used for hyper parameter optimization and the optimal model configuration.

We trained the LSTM model for 250 epochs with an early stopping of 30 epochs. The epoch with the highest validation performance was selected according to the area under the curve (AUC) of the receiver operating characteristics (ROC) \citep{Metz1978}.

All deep learning was performed in PyTorch \citep{Paszke2019} and the code is accessible on GitHub\footnote{\url{https://github.com/WFSRDataScience/XAI4TTX}}.

\subsubsection{Evaluation}\label{eval}
For evaluation, we calculated the ROC curve on the validation set \citep{Metz1978}. The validation ROC curve was bootstrapped 10,000 times, yielding 95\% confidence intervals (95\% CI). The bootstrapping entails sampling from the validation set with replacement, then calculating new sensitivity, i.e., true positive rate, and specificity, i.e., true negative rate, combinations for different thresholds. From the ROC curve, we calculated the AUC and specificity at a sensitivity of 90\%. Afterwards, we calculated the ROC for the test set and evaluated whether the AUC and specificity fall within the bootstrapped range. If the test AUC and specificity fall within the bootstrapped range, it represents a well-fitted model. Furthermore, we also evaluated the test set on the threshold associated with a validation sensitivity of 90\%. For the threshold evaluation, we used sensitivity and specificity. 

\subsection{Explainable AI}
We used the XAI method SHAP \citep{Lundberg2017} to explain how the LSTM model came to its results. SHAP provides explanations for each feature of the LSTM prediction. SHAP provides both local explanations, i.e., per sample, and global explanations, i.e., for the entire dataset, therefore unfolding the relationships between hydrological and meteorological features and TTX contamination in bivalve mollusks. The SHAP results were tested for significance with the Mann-Whitney U Test.

\subsection{Sensitivity analysis}
We performed a sensitivity analysis to estimate the adaptability of the created model. For the sensitivity analysis, we tested the LSTM model’s prediction performance for TTX contamination above the LL instead of above the AL. The LSTM model was not retrained for this task, and instead only data from 2022 (validation) and 2023 (test) were used. The same method as described in \ref{eval} was used for the sensitivity analysis.

\section{Funding}
This project has received funding from the European Union’s HORIZON-CL6-2022  research and Innovation programme under grant agreement N◦101084201.
Furthermore, Funding for this research has also been provided by the Dutch Ministry of Agriculture, Fisheries, Food Security and Nature (KB-50-005-008).


\begin{appendices}

\section{Supplementary Figures and Tables}\label{suppple}

\begin{figure}[h]
\centering
\includegraphics[width=0.6\textwidth]{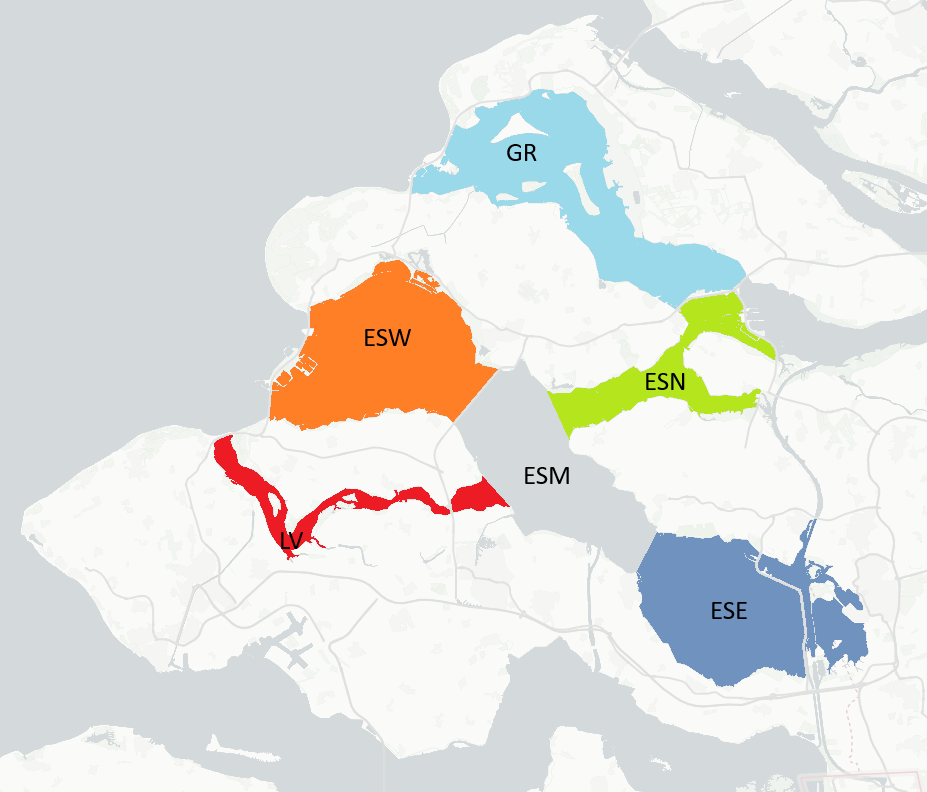}
\caption{
Measurement regions in Zeeland. The regions are the Eastern Scheldt east (ESE), north (ESN), west (ESW), middle (ESM), Lake Grevelingen (GR), and Lake Veere (LV). Tetrodotoxin (TTX) was detected in a total of 222 samples and no TTX in 934; the number of samples for each region are ESE (N=222), ESN (N=204), ESW (N=207), ESM (N=208), GR (N=167), VM (N=148).
}\label{REGIONS}
\end{figure}

\begin{table}[h]
\caption{Used hyperparameters for model development. The optimal and final parameters are selected based on validation Area Under the Curve (AUC) score.}\label{PARAMS}
\begin{tabular}{@{}lll@{}}
\toprule
Hyperparameter & Search space & Final model\\
\midrule
LSTM layers & 1-5 & 3 \\
Batch size & 16/32/64 & 32 \\
Learning rate & $10^{-3}$ - $10^{-5}$ & $5 \cdot 10^{-4}$ \\
Dropout & 0.2-0.5 & 0.3 \\
Hidden dimension & 64-128-256-512 & 256 \\
Learning rate decay & NA & 0.1 \\
Step size & NA & 30 \\
\botrule
\end{tabular}
\end{table}

\begin{sidewaystable}[t]
\caption{Mean (standard deviation) of the 35-day average of the meteorological and hydrological features.}\label{DATASIZE}
\centering
\small
\begin{tabular}{@{}lllll@{}}
\toprule
\textbf{Feature} & \textbf{ALL (N = 1,156)} & \textbf{Above AL (N = 75)} & \textbf{Above LL (N = 44)} & \textbf{No TTX (N = 934)} \\
\midrule
Week number – (1--52) & 27.6 ($\pm$ 10.4) & 24 ($\pm$ 2.08) & 23.4 ($\pm$ 1.56) & 28 ($\pm$ 10.4) \\
Sunrise – hour & 5.21 ($\pm$ 1.33) & 3.86 ($\pm$ 0.26) & 3.79 ($\pm$ 0.13) & 5.30 ($\pm$ 1.33) \\
Sunset – hour & 18.78 ($\pm$ 1.41) & 20.24 ($\pm$ 0.25) & 20.29 ($\pm$ 0.10) & 18.68 ($\pm$ 1.41) \\
Sun hours – number of hours & 13.57 ($\pm$ 2.74) & 16.38 ($\pm$ 0.50) & 16.50 ($\pm$ 0.23) & 13.37 ($\pm$ 2.72) \\
Global radiation – J/cm\textsuperscript{2} & 69.89 ($\pm$ 24.11) & 88.15 ($\pm$ 19.17) & 90.20 ($\pm$ 19.06) & 68.62 ($\pm$ 23.91) \\
Temperature – \celsius & 15.06 ($\pm$ 4.8) & 17.09 ($\pm$ 1.26) & 16.93 ($\pm$ 1.14) & 14.92 ($\pm$ 4.93) \\
Max temperature – \celsius & 18.32 ($\pm$ 5.39) & 20.88 ($\pm$ 1.54) & 20.79 ($\pm$ 1.46) & 18.14 ($\pm$ 5.52) \\
Min temperature – \celsius & 12.18 ($\pm$ 4.41) & 13.83 ($\pm$ 1.15) & 13.62 ($\pm$ 0.95) & 12.06 ($\pm$ 4.52) \\
Water temperature – \celsius & 16.18 ($\pm$ 5.19) & 18.24 ($\pm$ 1.49) & 18.00 ($\pm$ 1.25) & 16.04 ($\pm$ 5.32) \\
Water height – cm & 2.04 ($\pm$ 6.93) & 0.23 ($\pm$ 5.21) & -1.18 ($\pm$ 4.78) & 2.16 ($\pm$ 7.02) \\
Water height calculated – cm & 1.82 ($\pm$ 5.13) & -2.38 ($\pm$ 2.53) & -3.38 ($\pm$ 2.16) & 2.11 ($\pm$ 5.14) \\
Mean wind speed – m/s & 3.51 ($\pm$ 0.48) & 3.21 ($\pm$ 0.18) & 3.18 ($\pm$ 0.18) & 3.53 ($\pm$ 0.49) \\
Average wind direction – \degree & 174.9 ($\pm$ 63.7) & 194.7 ($\pm$ 63.9) & 194.4 ($\pm$ 66.4) & 173.6 ($\pm$ 63.5) \\
pH & 8.20 ($\pm$ 0.15) & 8.24 ($\pm$ 0.09) & 8.24 ($\pm$ 0.08) & 8.19 ($\pm$ 0.15) \\
Oxygen concentration – mg/L & 8.77 ($\pm$ 1.33) & 8.41 ($\pm$ 0.69) & 8.48 ($\pm$ 0.57) & 8.80 ($\pm$ 1.36) \\
Oxygen saturation – \% & 101.2 ($\pm$ 7.16) & 101.9 ($\pm$ 5.83) & 102.2 ($\pm$ 5.25) & 101.1 ($\pm$ 7.24) \\
Chlorophyll concentration – \micro g/L & 4.98 ($\pm$ 3.27) & 3.93 ($\pm$ 2.11) & 3.20 ($\pm$ 1.21) & 5.05 ($\pm$ 3.32) \\
Pheophytin concentration – \micro g/L & 0.12 ($\pm$ 0.06) & 0.12 ($\pm$ 0.06) & 0.10 ($\pm$ 0.05) & 0.12 ($\pm$ 0.06) \\
Air pressure – hPa & 1016 ($\pm$ 5) & 1014 ($\pm$ 5) & 1014 ($\pm$ 5) & 1016 ($\pm$ 6) \\
Chlorosity – g/L & 13.05 ($\pm$ 4.80) & 15.35 ($\pm$ 2.35) & 15.69 ($\pm$ 1.74) & 12.89 ($\pm$ 4.89) \\
Chloride concentration – mg/L & 9.95 ($\pm$ 7.52) & 7.49 ($\pm$ 7.82) & 8.66 ($\pm$ 7.91) & 10.12 ($\pm$ 7.47) \\
Conductivity – mS/m & 2971 ($\pm$ 1141) & 3675 ($\pm$ 597) & 3741 ($\pm$ 450) & 2922 ($\pm$ 1153) \\
Salinity – g/L & 29.74 ($\pm$ 1.82) & 29.31 ($\pm$ 1.37) & 29.54 ($\pm$ 0.97) & 29.77 ($\pm$ 1.85) \\
\bottomrule
\end{tabular}
\end{sidewaystable}

\begin{table}[t]
\caption{Number of samples in the train, validation, and test sets. The total number of samples is the sum of the samples with ‘no TTX’ and ‘above LOD’. With an additional distinction for samples with tetrodotoxin (TTX) above Action Limit (AL), Limit of Detection (LOD), and Legal Limit (LL). For the LOD and LL, no training was carried out.}\label{DATADIST}
\begin{tabular}{@{}lllllll@{}}
\toprule
 & Years & Total & No TTX & Above LOD & Above AL & Above LL\\
\midrule
Train & 2016-2021 & 857 & 726 (85\%) & 131 (15\%) & 47 (5\%) & 25 (3\%) \\
Validation & 2022 & 155 & 118 (76\%) & 37 (24\%) & 13 (8\%) & 9 (6\%) \\
Test & 2023 & 144 & 90 (63\%) & 54 (38\%) & 15 (10\%) & 10 (7\%) \\
Total & 2016-2023 & 1156 & 934 (81\%) & 222 (19\%) & 75 (6\%) & 44 (4\%) \\
\botrule
\end{tabular}
\end{table}

\end{appendices}
\end{document}